\documentclass[letterpaper, 10 pt, conference]{ieeeconf}  % Comment this line out if you need a4paper

\IEEEoverridecommandlockouts                              % This command is only needed if you want to use the \thanks command
\usepackage{times}
\usepackage{epsfig}
\usepackage{graphicx}
\usepackage{amsmath,amssymb}
\usepackage{multirow}
\usepackage{cite}
\usepackage{url}
\usepackage{xcolor}
\usepackage[normalem]{ulem}
\usepackage{siunitx}
\usepackage{stfloats}
\usepackage{booktabs}
\usepackage{hyperref}
\usepackage{cleveref}

\title{\LARGE \bf
Learning Energy-Efficient Air--Ground Actuation for Hybrid Robots \\on Stair-Like Terrain
}

\author{Jiaxing Li$^{*1}$, Ishaan Bhimwal$^{*1,2}$, Wen Tian$^{1}$, Xinhang Xu$^{3}$, Junbin Yuan$^{1}$, Yuxin Guo$^{1}$, \\Sebastian Scherer$^{1}$, Muqing Cao$^{4,1}$
\thanks{\raggedright $^{*}$Equal contribution. $^{1}$Carnegie Mellon University, US. $^{2}$Technische Hochschule W\"{u}rzburg-Schweinfurt, Germany. $^{3}$Nanyang Technological University, Singapore. $^{4}$Singapore University of Technology and Design, Singapore.\newline
Emails: {\tt \{\href{mailto:jiaxingl@andrew.cmu.edu}{jiaxingl},\allowbreak \href{mailto:ibhimwal@andrew.cmu.edu}{ibhimwal},\allowbreak \href{mailto:went@andrew.cmu.edu}{went},\allowbreak \href{mailto:junbiny@andrew.cmu.edu}{junbiny},\allowbreak \href{mailto:basti@andrew.cmu.edu}{basti},\allowbreak \href{mailto:muqingc@andrew.cmu.edu}{muqingc}\}@andrew.cmu.edu}, {\tt \href{mailto:xinhang.xu@ntu.edu.sg}{xinhang.xu@ntu.edu.sg}}
}}

\begin{document}
\maketitle
\thispagestyle{empty}
\pagestyle{empty}

%===========================================================

\begin{abstract}

Hybrid aerial--ground robots can use thrust to cross obstacles that impede
wheel-driven motion, but deciding how much thrust to apply during contact remains
challenging. We present an energy-aware reinforcement learning framework that
jointly commands wheels, tilt servos and propellers through a single continuous
policy, without prescribing locomotion modes. Hardware-calibrated power models
penalise estimated electrical energy, while a terrain curriculum and a terminal
reward for upright, settled arrivals support learning of thrust-assisted
climbing. In simulation, continuous thrust allocation improves single-step
clearance over fixed-thrust and mode-switching baselines as steps become taller.
 At the wheel-radius step height, it draws approximately
\SI{27}{\percent} less mean power than the best fixed allocation. An energy-weight ablation shows the accompanying trade-off between efficiency and reliability. On a
physical DoubleBee prototype, the unchanged network with a deployment interface
clears two \SI{6}{\centi\metre} steps in 8 of 10 trials. Additional hardware tests
show both the potential for transfer to different terrain geometries and the
remaining limitations in heading control, contact robustness and recovery.

\end{abstract}

%===========================================================

\section{Introduction}
\label{sec:intro}

Hybrid aerial--ground robots combine economical ground travel with the ability to lift over obstacles that wheels cannot cross. Many platforms exploit this complementarity, including protective rolling cages~\cite{kalantari2014hytaq}, passive or driven wheels mounted on multirotors~\cite{qin2020passivewheel,pan2023skywalker,kalantari2020drivocopter}, reconfigurable airframes that share mechanisms between modes~\cite{gefen2022fstar2} and morphing robots that repurpose their appendages~\cite{sihite2023m4}. The common idea is to drive where the ground allows and to use aerial actuation where ground contact alone is not sufficient.

A step, however, presents an allocation problem rather than a simple choice between driving and flying. Thrust can support part of the robot's weight and help the wheels pass the edge, but it also changes how strongly the wheels are loaded and how the body is oriented. With too little thrust, the robot lacks the authority to climb or to stay balanced. With too much, it spends unnecessary electrical energy and reduces the traction available to the wheels. The relevant control decision is therefore when to apply thrust, how much to apply and in which direction, while the wheels continue to drive.

\begin{figure}
    \centering
    \includegraphics[width=1\linewidth]{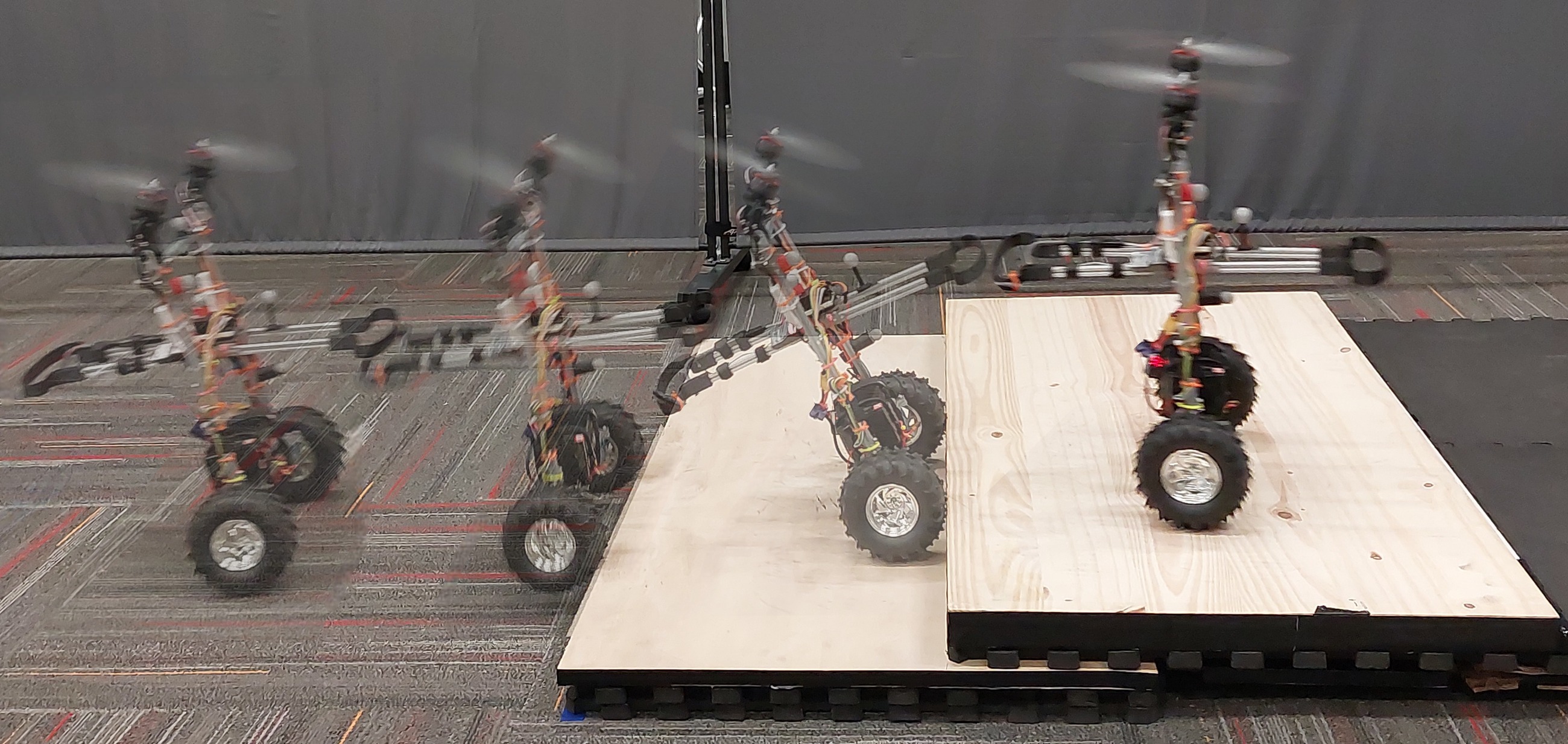}
    \caption{Thrust-assisted climbing on DoubleBee. Four exposures show the robot approaching and traversing two \SI{6}{\centi\metre} steps spaced \SI{40}{\centi\metre} apart. The wheels provide forward travel while thrust assists passage over each edge, and the robot covers \SI{1.3}{\metre} in \SI{2.9}{\second} before settling upright on the top step.}
    \label{fig:overview}
\end{figure}

Hybrid locomotion is commonly organised into discrete modes, with planners or switching rules selecting among them~\cite{gefen2022fstar2,fan2019autonomoushybrid,suh2020energyefficient,sihite2023m4}. Such architectures can include an assisted-ground mode in which thrust and wheels act together, but the allocation of effort within that mode and its timing at the edge still have to be designed by hand. This low-level allocation is the focus of this paper.

Reinforcement learning has enabled contact-rich locomotion through large-scale simulation and domain randomisation~\cite{tobin2017domainrand,rudin2021walkminutes}. Learned policies climb stairs without vision on bipeds~\cite{siekmann2021blindstairs} and on legged and wheeled-legged robots~\cite{chamorro2024blindstair}, and two-wheeled platforms have been extended with articulated legs~\cite{klemm2019ascento}. Other work has learned transitions between walking and flying with adversarial motion priors~\cite{lerario2024walkflyamp} and compared learned and model-based control for quadrotor morpho-transition~\cite{mandralis2025morpho}. These methods learn how to move and when to change locomotion style, but they do not address how effort should be divided between propulsion sources with different energetic costs.

Energy has also been used directly as a learning objective. Minimising energy consumption causes distinct gaits to emerge in legged robots~\cite{fu2021emergentgaits}, and optimising cost of transport produces learned gait transitions~\cite{yang2021gaittransitions}. Hybrid aerial--ground robots add a different kind of coupling. Wheel and propeller effort have very different electrical costs, and their mechanical effects change together when the robot meets a step, since thrust reduces wheel loading while tilting the propellers redirects thrust and pitches the body at the same time. The choice also depends on terrain geometry rather than on a commanded speed, so the policy must perceive the step before deciding how to cross it. A generic penalty on action magnitude treats all actuators alike and therefore does not represent the difference in cost.

We study this problem on DoubleBee~\cite{cao2023doublebee}, a two-wheeled balancing robot with tilting propellers. A policy trained in Isaac Lab~\cite{mittal2025isaaclab} receives a local height scan and the robot state and continuously commands the wheels, tilt servos and propellers at every control step, without a mode variable. The reward combines an estimate of electrical energy, computed from hardware-calibrated thrust and power models, with task and posture terms. Energy minimisation alone is not expected to discover successful climbs, so these terms guide the policy toward thrust-assisted climbing. We compare the policy with fixed-thrust and mode-switching baselines under a matched evaluation protocol and test it on the physical robot. Our contributions can be summarised as follows:

\begin{itemize}
    \item We present an energy-aware reinforcement learning framework for joint wheel, servo and propeller control, in which a single continuous policy is trained with hardware-calibrated power models and a terminal reward that requires an upright, settled arrival.
    \item We compare continuous thrust allocation with fixed-thrust and mode-switching
    baselines under a matched evaluation protocol, separating gains in single-step
    clearance from reductions in power and energy.
    \item We evaluate the learned policy on hardware with an explicit deployment interface, including repeated two-step climbs, tests on additional terrain geometries and an analysis of contact, heading and recovery failures.
\end{itemize}

%===========================================================

\section{System Overview}
\label{sec:system}

\subsection{Platform and Actuation}
\label{subsec:dynamics}

\begin{figure}[t]
    \centering
    \includegraphics[width=0.50\linewidth]{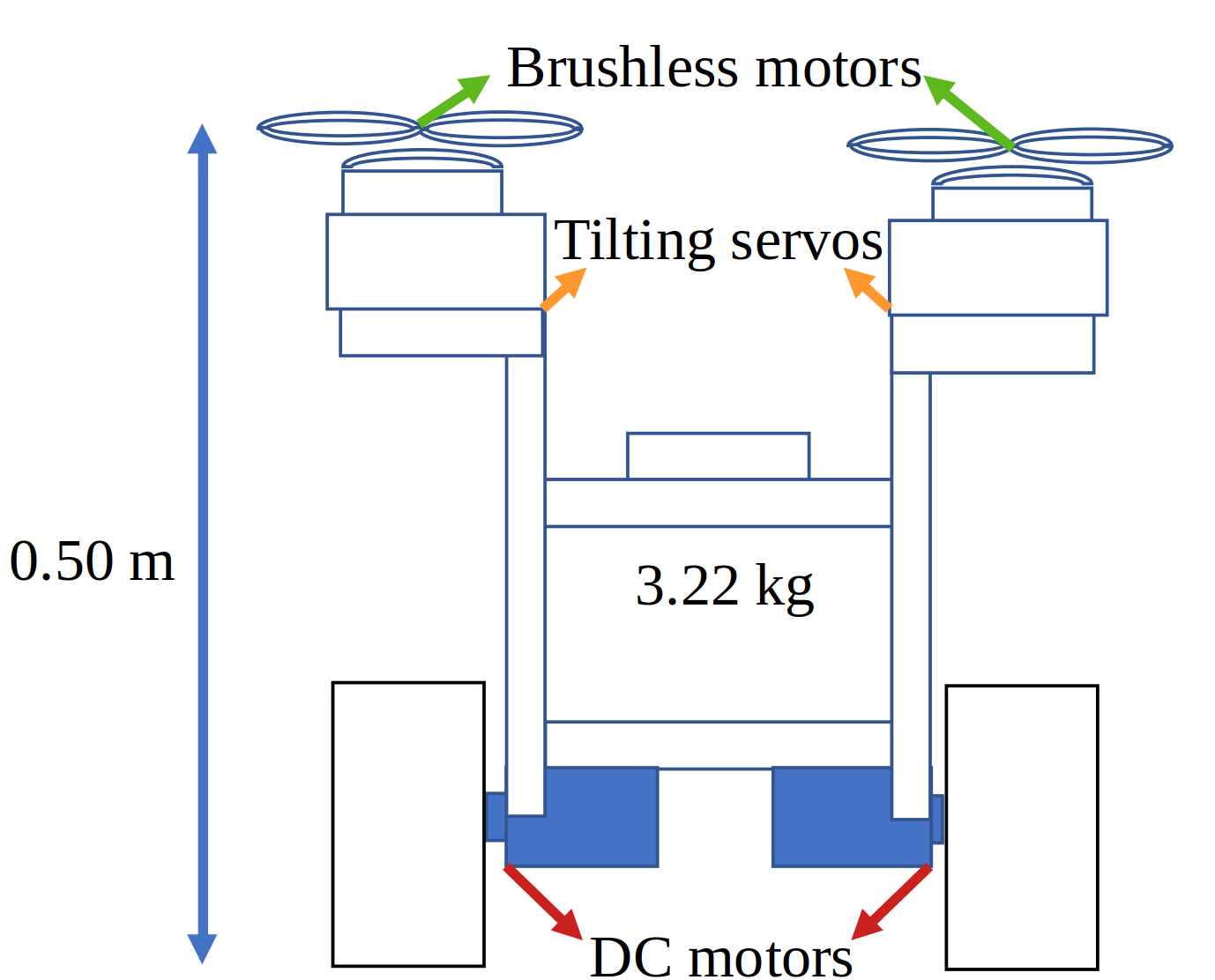}\hfill
    \includegraphics[width=0.43\linewidth]{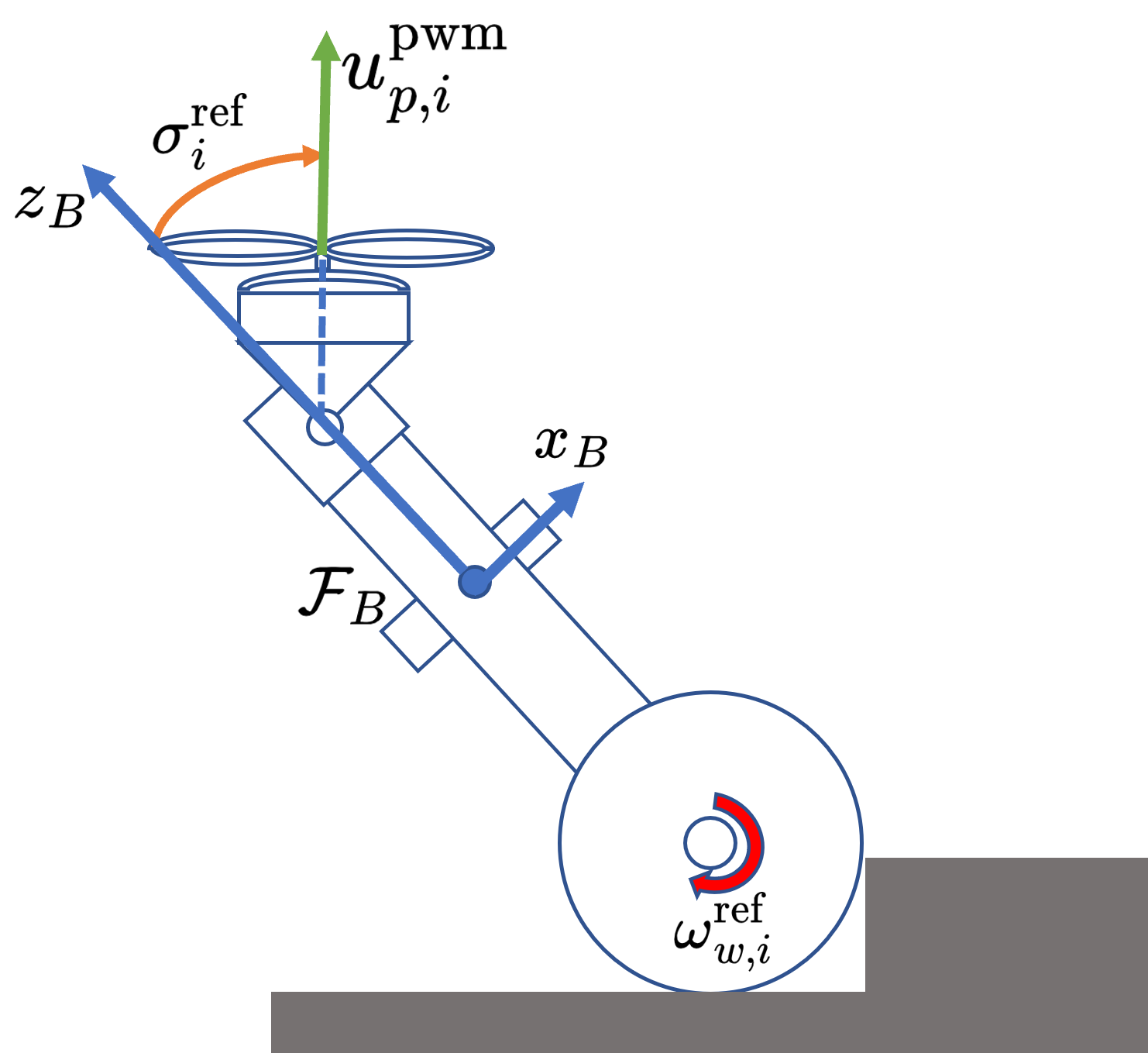}
    \caption{DoubleBee and its actuators. The policy uses a shared propeller
    command, a shared servo command, and common and differential wheel commands.
    This four-dimensional action is mapped to the six low-level channels (\Cref{subsec:obs_action}).}
    \label{fig:actuators}
\end{figure}

We build on the DoubleBee hybrid aerial--ground robot \cite{cao2023doublebee},
shown in \Cref{fig:actuators}. The robot balances on two independently driven
wheels, much like a two-wheeled inverted pendulum, and carries two propeller
units above the wheel axle. Each propeller unit is mounted on a servo that tilts
its thrust forward or backward. At the lowest level, the robot accepts six
commands, namely a speed reference for each wheel, an angle reference for each
servo and a PWM command for each propeller. We describe the robot in the body
frame $\mathcal{F}_B$, with $x_B$ pointing forward, $y_B$ lateral and $z_B$ upward
along the body, and denote the pitch angle about $y_B$ by $\vartheta$. Each
propeller $i$ produces a thrust of magnitude $f_{p,i}$ along its own axis, and the
servo angle $\sigma_i$ tilts this axis about $y_B$, giving the body-frame force
\begin{equation}
\mathbf{f}^{B}_{p,i} = f_{p,i}\begin{bmatrix}\sin\sigma_i & 0 & \cos\sigma_i\end{bmatrix}^{\top},
\quad f_{p,i}\ge 0,\ i\in\{1,2\},
\label{eq:thrust}
\end{equation}
so that a positive $\sigma_i$ tilts the thrust toward $x_B$.

Thrust acts on the body at any attitude, whether or not the wheels are on the
ground. Its magnitude sets how much of the robot's weight the propellers support,
and its tilt sets both the direction of the force and its moment about the body.
Wheel torque, in contrast, acts only through wheel--terrain contact, which is
limited by friction and normal load. The two are coupled, because vertical thrust
reduces wheel load and therefore traction, so the inputs must be coordinated
rather than chosen independently.

\subsection{Why Thrust Helps at a Step}
\label{subsec:balance}

Two static estimates explain why thrust helps at a step. The first concerns
lifting the wheels over the edge. With both wheels at a rigid step edge and a
vertical thrust $f_p=f_{p,1}+f_{p,2}$, a step of height $0<h\le r$ places the edge
a horizontal distance $d=\sqrt{2rh-h^{2}}$ from the axle of a wheel of radius $r$,
about which the wheel must rotate to climb. Sharing the load equally, each wheel
needs approximately
\begin{equation}
\tau_{\mathrm{req}} \approx \tfrac12\,(mg-f_p)\,d.
\label{eq:wheel_requirement}
\end{equation}
Without thrust this exceeds the \SI{0.51}{\newton\metre} motor limit for all but
very small steps ($m=\SI{3.22}{\kilo\gram}$, $r=\SI{0.06}{\metre}$). At $h=r$,
about \SI{14.6}{\newton} of thrust, or \SI{46}{\percent} of body weight, brings it
within the limit, so partial unloading suffices and the robot does not need to
hover. Above $h=r$ the wheel meets the face at or above its widest point, where
forward torque presses it against the face rather than lifting it, so we use
$h_{\max}=r$ as the reference step height. The estimate is quasi-static and
ignores momentum, wheel slip and the reaction torque that wheel acceleration
applies to the body.

The second estimate concerns balance. Wheels pressed against a step cannot move
freely under the body to correct a lean, so thrust offers another way to stay
upright. Thrust kept vertical while the body leans by $\vartheta$ acts
$L_p=\SI{0.44}{\metre}$ above the axle and gives a restoring moment
$f_pL_p\sin\vartheta$, against gravity's $mgL_{\mathrm{com}}\sin\vartheta$ at the
centre of mass, $L_{\mathrm{com}}=\SI{0.14}{\metre}$ above the axle. The net
moment restores the robot toward upright whenever
\begin{equation}
f_p > \frac{mg\,L_{\mathrm{com}}}{L_p},
\label{eq:stability_threshold}
\end{equation}
which is about \SI{10}{\newton}, or one third of body weight. This condition does
not slow the rotation, so the wheels and changes in thrust must still damp the
motion. How much thrust to apply for lifting and for balancing, and when to apply
it during approach, contact and arrival, is what the policy has to learn.

\section{Energy-Aware Policy Learning}
\label{sec:task_learning}

% The policy must make progress, remain recoverable during contact and avoid unnecessary propulsion energy.

An ideal policy should progress toward the target by driving and lifting the robot over discontinuous terrain, while remaining recoverable during contact and avoiding unnecessary propulsion energy. We first define its task and information, then the continuous action and reward. Actuator calibration and the deployment interface are described in \Cref{sec:sim2real}, and the design choices are evaluated in \Cref{sec:results}.

\subsection{Task and Terrain Curriculum}
\label{subsec:sim_setup}
\label{subsec:task}

\begin{figure}[t]
    \centering
    \includegraphics[width=\linewidth]{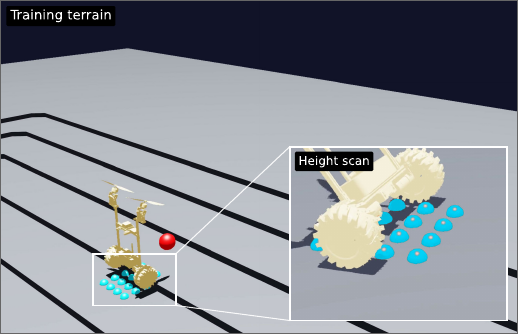}
    \caption{Training terrain and the policy's view of it. Each
    $10\times10$\,\si{\meter} tile is a procedurally generated inverted-pyramid
    staircase with a \SI{0.40}{\meter} tread and step heights from
    \SIrange{0.03}{0.08}{\meter}, instantiated at five difficulty levels with five
    tiles each. Episodes start on a flat
    patch at the base of the staircase, with the goal (red) sampled
    \SIrange{1.5}{3.2}{\meter} ahead and \SIrange{0.03}{0.20}{\meter} above it.
    \textit{Inset:} the height scan (cyan), a $4\times4$ grid of rays at
    \SI{0.07}{\meter} spacing over a \SI{0.21}{\meter} square centred on the base
    and aligned with its heading.} % , shown from the side.
    \label{fig:terrain}
\end{figure}

Each episode starts on a flat patch below a procedurally generated
inverted-pyramid staircase and asks the robot to reach an elevated goal
(\Cref{fig:terrain}). Step height varies while tread length remains fixed.
Depending on the sampled height and goal, reaching the target requires an
ascent of one to six steps. The task specifies a destination rather than a
reference velocity, leaving the policy to choose its approach and climbing speed.

The terrain curriculum promotes an environment after three consecutive successes
and demotes it after five consecutive failures. The different thresholds are
intended to retain exposure to harder terrain when early success is sparse,
without preventing a return to an easier level.

The goal command, which provides a direction and a bounded distance cue, is
\begin{equation}
  \mathbf{cmd}_t=[d_x^B,d_y^B,0,\bar\rho]^\top,
  \qquad \bar\rho=\min(\rho/d_{\max},1).
  \label{eq:goal_cmd}
\end{equation}
Here $[d_x^B,d_y^B]^\top$ is the unit horizontal direction to the goal in the
body frame, $\rho$ is horizontal distance and $d_{\max}=\SI{2}{\metre}$.
The policy can therefore use distance to distinguish approach from arrival
without receiving a prescribed speed or a locomotion-mode label. The third entry is a yaw-rate slot in the command format and is held at zero.

% The reserved yaw-rate slot is held at zero, because the wheel differential is limited to a small trim (\Cref{subsec:obs_action}).

\subsection{Observation and Continuous Action}
\label{subsec:obs_action}

The 41-dimensional observation combines robot state, actuator state, the goal
command, local terrain and the previous action (\Cref{tab:observations}). Wheel-contact indicators are max-pooled over a short force history. The height scan is the only terrain input, and its short look-ahead of about \SI{0.10}{\metre} makes coordination at contact important, since the policy cannot plan a long sequence of steps from the observation alone.

% Binary wheel-contact

The policy outputs
\begin{equation}
  \mathbf a_t=[a^w_{\mathrm{com}},a^w_{\mathrm{diff}},a^\sigma,a^p]^\top
  \in[-1,1]^4.
  \label{eq:action}
\end{equation}
The common wheel component primarily drives translation and contributes to
balancing, while the smaller differential component provides yaw trim. Their scales
are \SI{23.6}{\radian\per\second} and \SI{4}{\radian\per\second}, respectively.
The paired servos share the angle reference $(\pi/4)a^\sigma$, and the paired
propellers share the spin reference $320(1+a^p)$\,\si{\radian\per\second}.
The latter passes through the simulated actuator dynamics before conversion to
thrust.

% Sharing the propeller and servo commands restricts the policy to the symmetric
% actuation useful for predominantly sagittal climbing. It also removes opposed
% commands within a pair, which were problematic during transfer. The wheels
% retain independent common and differential components. This is a reduced
% continuous action space, not a mode switch, since all four outputs are available at
% every control step.

Sharing the propeller and servo commands restricts the policy to the symmetric actuation useful for predominantly sagittal climbing. The wheels retain independent common and differential components, and all four outputs remain available at every control step.

% \begin{table}[t]
% \centering
% \caption{Policy observation. Paired actuator quantities contain one value per
% left/right unit.}
% \label{tab:observations}
% \footnotesize
% \begin{tabular}{@{}l c p{0.55\columnwidth}@{}}
% \toprule
% Signal & Dim. & Information \\
% \midrule
% $\boldsymbol\omega_w$ & 2 & Wheel angular velocities \\
% $\boldsymbol\sigma$ & 2 & Servo tilt angles \\
% $\boldsymbol\omega_p$ & 2 & Propeller spin rates \\
% $\mathbf v^B$ & 3 & Body-frame linear velocity \\
% $\boldsymbol\omega^B$ & 3 & Body-frame angular velocity \\
% $\hat{\mathbf g}$ & 3 & Projected gravity (attitude) \\
% $\mathbf c^w$ & 2 & Binary wheel-contact indicators \\
% $\mathbf{cmd}$ & 4 & Goal direction and range, \eqref{eq:goal_cmd} \\
% $\Delta\mathbf h$ & 16 & Local $4\times4$ height scan \\
% $\mathbf a_{t-1}$ & 4 & Previous policy action \\
% \midrule
% Total & 41 & \\
% \bottomrule
% \end{tabular}
% \end{table}

\begin{table}[t]
\centering
\caption{Policy observation. Paired actuator quantities contain one value per
left/right unit.}
\label{tab:observations}
\footnotesize
\renewcommand{\arraystretch}{1.2}
\begin{tabular*}{\columnwidth}{@{\extracolsep{\fill}}l c l@{}}
\toprule
Signal & Dim. & Information \\
\midrule
$\boldsymbol\omega_w$ & 2 & Wheel angular velocities \\
$\boldsymbol\sigma$ & 2 & Servo tilt angles \\
$\boldsymbol\omega_p$ & 2 & Propeller spin rates \\ \addlinespace[2pt]
$\mathbf v^B$ & 3 & Body-frame linear velocity \\
$\boldsymbol\omega^B$ & 3 & Body-frame angular velocity \\
$\hat{\mathbf g}$ & 3 & Projected gravity (attitude) \\
$\mathbf c^w$ & 2 & Binary wheel-contact indicators \\ \addlinespace[2pt]
$\mathbf{cmd}$ & 4 & Goal direction and range, \eqref{eq:goal_cmd} \\
$\Delta\mathbf h$ & 16 & Local $4\times4$ height scan \\
$\mathbf a_{t-1}$ & 4 & Previous policy action \\
\midrule
Total & 41 & \\
\bottomrule
\end{tabular*}
\end{table}

\subsection{Reward and Successful Completion}
\label{subsec:reward}

A distance reward alone does not specify how to survive the wheel--step contact,
whereas posture rewards can favour standing still. We therefore combine task
progress, mechanically motivated propeller engagement, stability, propulsion
energy and terminal outcomes in the reward
\begin{equation}
  r_t=w_{\mathrm{task}}r_t^{\mathrm{task}}\!+\!w_{\mathrm{prop}}r_t^{\mathrm{prop}}
      \!+\!w_{\mathrm{stab}}r_t^{\mathrm{stab}}\!+\!w_Ep_t^E\!+\!r_t^{\mathrm{term}}.
  \label{eq:reward}
\end{equation}
The principal component terms and their weights are given in
\Cref{tab:rewards}. The engagement terms encourage useful support and thrust
direction, but they do not prescribe an actuator sequence. Consequently, the policy learns continuous allocation under shaped task constraints, rather than from electrical cost alone.

% \begin{table}[t]
% \centering
% \caption{Principal reward components and weights. Group names correspond to
% \eqref{eq:reward}; terminal values include the listed outcome weights.}
% \label{tab:rewards}
% \footnotesize
% \begin{tabular}{@{}p{0.78\columnwidth} r@{}}
% \toprule
% Component & Weight \\
% \midrule
% \multicolumn{2}{@{}l}{\emph{Task progress}}\\
% Goal progress; height gain; forward progress; settled climb & 2 each\\
% Per-step reward within $60^\circ$ of upright & 0.5\\
% \multicolumn{2}{@{}l}{\emph{Propeller engagement}}\\
% Thrust under lean & 1.5\\
% Keeping thrust vertical; upward thrust at a step & 2 each\\
% Vertical thrust at 0.7 of body weight & 1.5\\
% \multicolumn{2}{@{}l}{\emph{Penalties}}\\
% Tilt beyond a $3.6^\circ$ dead-zone & 4.5\\
% Electrical energy, \eqref{eq:energy_penalty} & 4\\
% Downward thrust; roll/pitch tilt & 3; 1\\
% Stalling near the target & 6\\
% \multicolumn{2}{@{}l}{\emph{Terminal outcomes}}\\
% Goal reward, \eqref{eq:terminal_goal} & $w_g=10$\\
% Propeller collision; leaving the arena & 3; 1\\
% \bottomrule
% \end{tabular}
% \end{table}

\begin{table}[t]
\centering
\caption{Principal reward components and weights. Group names correspond to
% \eqref{eq:reward}, and terminal values include the listed outcome weights.}
\eqref{eq:reward}. Terminal weights multiply a base value of 100 for the goal
reward and $-10$ for each failure.}
\label{tab:rewards}
\footnotesize
\begin{tabular}{@{}p{0.78\columnwidth} r@{}}
\toprule
Component & Weight \\
\midrule
\multicolumn{2}{@{}l}{\emph{Task progress}}\\
Goal progress, height gain, forward progress, settled climb & 2 each\\
Per-step reward within $60^\circ$ of upright & 0.5\\
\multicolumn{2}{@{}l}{\emph{Propeller engagement}}\\
Vertical thrust while leaning, increasing with lean & 1.5\\
Keeping thrust vertical, upward thrust at a step & 2 each\\
Vertical thrust, normalised to 0.7 of body weight & 1.5\\
\multicolumn{2}{@{}l}{\emph{Penalties}}\\
Tilt outside a $3.6^\circ$ dead-zone & 4.5\\
Electrical energy, \eqref{eq:energy_penalty} & 4\\
Downward thrust, roll/pitch tilt & 3, 1\\
Stalling near the target & 6\\
\multicolumn{2}{@{}l}{\emph{Terminal outcomes}}\\
Goal reward, \eqref{eq:terminal_goal} & $w_g=10$\\
Propeller collision, leaving the arena & 3, 1\\
\bottomrule
\end{tabular}
\end{table}

\subsubsection{Electrical-energy penalty}
An action norm treats equal-magnitude wheel and propeller commands as comparable,
even though their electrical costs differ. Instead, calibrated power functions
estimate the energy spent during one control interval as
\begin{align}
 E_t&=\Delta t\sum_{i=1}^{2}\left[
 P_{\mathrm{pwm}}(u^{\mathrm{pwm}}_{p,i})+
 P_{\mathrm{rpm}}\!\left(\frac{60}{2\pi}|\omega^{\mathrm{ref}}_{w,i}|\right)
 \right],\label{eq:energy_estimate}\\
 p_t^E&=\exp(-E_t/E_{\mathrm{ref}})-1,
 \qquad E_{\mathrm{ref}}=\SI{20}{\joule}.
 \label{eq:energy_penalty}
\end{align}
The exponential gives a bounded penalty that increases in magnitude with
expenditure. $P_{\mathrm{pwm}}$ models propeller power and $P_{\mathrm{rpm}}$
models wheel-drive power, and the conversion in \eqref{eq:energy_estimate} expresses
wheel speed in RPM. This is a model-based propulsion-energy objective, not a
direct measurement of total battery energy. In particular, the expression does
not explicitly include servo or auxiliary consumption. The wheel term is also a function of commanded speed, so it does not capture the current drawn by a stalled wheel. Hardware evaluation instead integrates measured voltage and current. We use $w_E=4$, and the choice of this weight and its effect on reliability are evaluated in \Cref{subsec:energy_ablation}.

% \subsubsection{Terminal reward}
% Reaching the goal ends the episode, and with it the per-step rewards the policy
% would have collected over the rest of the horizon, so a policy can prefer to
% linger near the goal rather than finish. We compensate for this forfeited return,
% \begin{equation}
%  r_t^{\mathrm{goal}}=w_gR_g+\lambda(T-t),
%  \qquad w_gR_g=1000,\quad\lambda=9,
%  \label{eq:terminal_goal}
% \end{equation}
% where $T$ and $t$ count control steps. The term $\lambda(T-t)$ pays for the steps
% given up by finishing early, with $\lambda$ set to the per-step return available
% without making progress, which is dominated by the posture and propeller terms
% rather than by survival. The compensation is approximate rather than an exact
% equality of discounted returns. Failures end the episode with a penalty of $-30$
% for propeller collision and $-10$ for leaving the arena.

\subsubsection{Terminal reward}
Reaching the goal ends the episode and forfeits the per-step reward the policy
would otherwise collect over the remaining horizon, so an uncompensated policy
can prefer to linger near the goal rather than finish. The goal reward therefore
includes a remaining-horizon term,
\begin{equation}
 r_t^{\mathrm{goal}}=w_gR_g+\lambda(T-t),
 \qquad w_gR_g=1000,\quad\lambda=9,
 \label{eq:terminal_goal}
\end{equation}
where $t$ counts control steps and $T=600$ is the episode horizon, twelve seconds at \SI{50}{\hertz}. We set $\lambda$ to the per-step return observed without progress, which is dominated by the posture and propeller terms rather than by survival alone. This is an approximate offset, not an exact equality of discounted returns. Failures end the episode with a penalty of $-30$ for propeller collision and $-10$ for leaving the arena.

% \subsubsection{Upright, settled arrival}
\subsubsection{Upright arrival}
The goal bonus is awarded only when four conditions hold simultaneously, namely
horizontal distance below \SI{0.25}{\metre}, elevation within
\SI{0.15}{\metre} of the target patch, a projected-gravity uprightness score above
0.85, and body angular-speed norm below \SI{3.5}{\radian\per\second}.
The conjunction prevents a fallen robot near the goal from receiving the same
terminal bonus as an upright arrival. It defines goal success in training and is
separate from the step-clearance metric used in the allocation experiments.
\Cref{subsec:success_metric} examines the effect of removing these conditions
when scoring the same trajectories.

%%%

\subsection{Training Procedure}
\label{subsec:training}

% We train with Truncated Quantile Critics (TQC)~\cite{kuznetsov2020tqc}, using MLP
% actor and critic networks. The critic ensemble represents return distributions,
% and dropping the highest quantiles limits optimistic value estimates. Experience is
% collected from 1024 parallel environments at \SI{50}{\hertz} and reused through
% replay. This makes successful contact transitions available for more than one
% update, which is useful when early successful climbs are sparse. Each of the 4000
% training iterations collects 24 steps in every environment and performs 24
% gradient updates, giving approximately 98 million transitions and \num{96000}
% updates on a single NVIDIA A100 GPU.

We train with Truncated Quantile Critics (TQC)~\cite{kuznetsov2020tqc}, using MLP
actor and critic networks. The critic ensemble represents return distributions,
and dropping the highest quantiles limits optimistic value estimates. Experience
is collected from 1024 parallel environments at \SI{50}{\hertz} and reused
through replay, so successful contact transitions contribute to more than one
update, which matters when early climbs are sparse.

Training has two phases. Pre-training starts from a balancing policy that does not climb and optimises the task, engagement and stability terms of \eqref{eq:reward} without the energy penalty for \num{1900} iterations, by which point thrust-assisted climbing has emerged. Energy fine-tuning then adds the electrical energy penalty  \eqref{eq:energy_penalty}  with $w_E = 4$ for a further \num{4000} iterations, and the pre-training checkpoint is the common starting point for the energy-weight ablation in \Cref{subsec:energy_ablation}. Each iteration collects 24 steps per environment and performs 24 gradient updates, so fine-tuning covers approximately 98 million transitions and \num{96000} updates, and both phases together approximately 145 million transitions, on a single NVIDIA A100 GPU.

\section{Sim-to-Real Transfer}
\label{sec:sim2real}
\subsection{Actuator and Power Models}
\label{subsec:sysid}

% \begin{table}[t]
% \centering
% \caption{Reported actuator envelopes. $T/W$ denotes total thrust divided by
% body weight, and wheel speed is limited to \SI{23.6}{\radian\per\second}.}
% \label{tab:actuators}
% \footnotesize
% \renewcommand{\arraystretch}{1.2}
% \begin{tabular}{@{}>{\raggedright}p{0.30\columnwidth} >{\raggedright}p{0.30\columnwidth} >{\raggedright\arraybackslash}p{\dimexpr0.40\columnwidth-4\tabcolsep\relax}@{}}
% \toprule
% Quantity & Simulation & Hardware \\
% \midrule
% Propeller limit & Approx.\ \SI{236}{\radian\per\second}\newline $T/W\approx0.50$
%  & \SI{1650}{\micro\second} cap, \SI{36.6}{\newton}\newline $T/W=1.16$\\ \addlinespace[2pt]
% Wheel acceleration & \SI{48.8}{\radian\per\second\squared}
%  & \SI{58}{\radian\per\second\squared}\\ \addlinespace[2pt]
% Servo rate & \SI{2}{\radian\per\second}
%  & \SI{10}{\radian\per\second}\\
% \bottomrule
% \end{tabular}
% \end{table}

The simulator uses a CAD-derived robot model with measured component masses and
CAD inertia tensors. Wheel and propeller joints use delayed velocity controllers,
and tilt servos use delayed position controllers. Saturation and rate limits matter
because the policy must coordinate support and attitude within the available
actuator response, rather than only command a desired final pose.

Propeller thrust is applied as an external force at each physics step. The
simulated spin rate is converted to an equivalent PWM value and then to thrust by
\begin{align}
 u^{\mathrm{pwm}}_{p,i}
 &=\operatorname{clip}(1000+1.3|\omega_{p,i}|,1000,1650)
   \,\si{\micro\second},\label{eq:omega_to_pwm}\\
 f_{p,i}&=\sum_{k=0}^{4}c_k(u^{\mathrm{pwm}}_{p,i})^{4-k}.
 \label{eq:pwm_to_thrust}
\end{align}
The fourth-order thrust fit has a calibration RMSE of
\SI{3.94e-3}{\newton}, and the propeller power fit used in
\eqref{eq:energy_estimate} has an RMSE of \SI{1.65}{\watt}. Using the same PWM
variable for thrust and power connects the mechanical action to its estimated
electrical cost.

The same calibration relates thrust to power. Over the measured range a power-law
fit gives $P\propto T^{1.49}$, close to the value of $1.5$ for an ideal
propeller, so a 30\% increase in propeller power yields about 19\% more thrust.
Increasing thrust is disproportionately costly in power, which is what the energy
term in \eqref{eq:energy_estimate} acts on.

%%%

\subsection{Randomisation and Latency}
\label{subsec:dr}

Randomisation exposes the policy to variations in loading, traction and actuator
response (\Cref{tab:randomisation}). Small intermittent pushes perturb balance
during the episode rather than only its initial condition. Observations are
delayed by one control step, and wheel commands have an additional transport
delay. These delays are important for a balancing robot, since the policy must act
on a state that has already evolved before its command reaches the wheels.

\begin{table}[t]
\centering
\caption{Randomisation and delay settings. Physical parameters are resampled at every reset, except actuator stiffness and damping, which are sampled once at startup.}
\label{tab:randomisation}
\footnotesize
\begin{tabular}{@{}p{0.42\columnwidth} p{0.52\columnwidth}@{}}
\toprule
Quantity & Perturbation or range\\
\midrule
Body mass / CoM location & $\pm5\%$ / $\pm\SI{1}{\centi\metre}$\\
Per-propeller thrust & $\pm20\%$\\
Static / dynamic friction & 0.8--1.2 / 0.7--1.0\\
Wheel/servo stiffness, damping & $\pm20\%$\\
Initial roll and pitch & $\pm\SI{0.2}{\radian}$\\
Push: linear / angular & $\pm\SI{0.05}{\metre\per\second}$ /
 $\pm\SI{0.25}{\radian\per\second}$\\
Push interval & \SIrange{3}{6}{\second}\\
Observation delay & \SI{20}{\milli\second}\\
Additional wheel delay & \SIrange{40}{80}{\milli\second}\\
\bottomrule
\end{tabular}
\end{table}

\subsection{Hardware Observation and Command Interface}
\label{subsec:deployment}

Observations are reconstructed at \SI{50}{\hertz} from motion-capture estimates
of base state, wheel encoders and a static terrain map. The local height scan is computed analytically from that map. Servo angle and propeller speed are reconstructed from the commanded action, and the wheel-contact indicators are fixed at one. 3-D printed bumpers are added on both sides of the chassis to protect the robot during falls.

% Identification reveals differences between the simulated and measured actuators.
% The simulated propeller saturates near \SI{236}{\radian\per\second} and holds
% total thrust to about half of body weight, where the bench calibration in
% \eqref{eq:pwm_to_thrust} reaches \SI{36.6}{\newton}, or 1.16 times body weight,
% at the \SI{1650}{\micro\second} cap. The simulated servo is rate limited to
% \SI{2}{\radian\per\second} against \SI{10}{\radian\per\second} measured. A
% calibrated command-to-thrust mapping keeps the relation between policy output and
% produced thrust consistent between simulation and hardware, so each propeller
% action retains the physical meaning it had in training. Wheel and servo commands
% are low-pass filtered, and the servo is rate limited to its simulated response.
% The propeller command is scaled and held above a minimum level. The wheel differential is held at zero, since on the robot this output produced unintended turning.

Identification reveals differences between the simulated and measured actuators.
The simulated propeller saturates near \SI{236}{\radian\per\second} and holds
total thrust to about half of body weight, whereas the bench calibration in
\eqref{eq:pwm_to_thrust} reaches \SI{36.6}{\newton}, or 1.16 times body weight,
at the \SI{1650}{\micro\second} cap. The simulated servo is rate limited to
\SI{2}{\radian\per\second} against \SI{10}{\radian\per\second} measured. A
calibrated command-to-thrust mapping relates policy output to produced thrust on
the robot, with the propeller command scaled and held above a minimum level.
Wheel and servo commands are low-pass filtered, and the servo is rate limited to
its simulated response. The wheel differential is held at zero, since on the
robot this output produced unintended turning.

%, so hardware trials drive straight toward the step.

% On the robot the propellers operate closer to the simulated value than to the
% calibrated cap. Battery power recorded during the trials stays far below the draw
% that sustained operation at the cap would require, and inverting the bench power
% fit places delivered thrust near \SI{17}{\newton}, an upper bound during
% transients since part of that power accelerates the rotors rather than producing
% thrust. We do not identify why the propellers fall short of the cap, and both
% speed-controller current limiting and incomplete spin-up are consistent with the
% measurements. The mismatch that remains is one of timing rather than magnitude,
% which is consistent with the jamming and stalled-lean failures in
% \Cref{subsubsec:hw_failures}.

This interface is fixed across the primary two-step trials. Parameters are
adjusted for the additional terrain configurations, and between trials on the
foam-step configuration. The network weights remain unchanged in all hardware
trials.

\section{Results and Discussion}
\label{sec:results}

We evaluate three questions, namely whether continuous thrust allocation improves
climbing, how its energy penalty changes the efficiency--reliability trade-off,
and which behaviours transfer to hardware. Simulation results aggregate the final
ten checkpoints of each run, evaluated for 200 episodes each with paired seeds for
terrain and goals, while the hardware experiments use the final checkpoint alone.
Each run is trained once, so seed-level variation is not measured. Throughout,
step clearance means climbing at least one step of the tested height, and it does
not imply the upright, settled goal completion defined in \Cref{subsec:reward}.
Power in watts and accumulated energy in joules are reported separately. Energy
comparisons conditioned on steps climbed are identified explicitly, since doing
less work can otherwise appear more efficient.

\subsection{Continuous Thrust versus Fixed Allocation}
\label{sec:accu_ablation}

\begin{figure}[t]
  \centering
  \includegraphics[width=\linewidth]{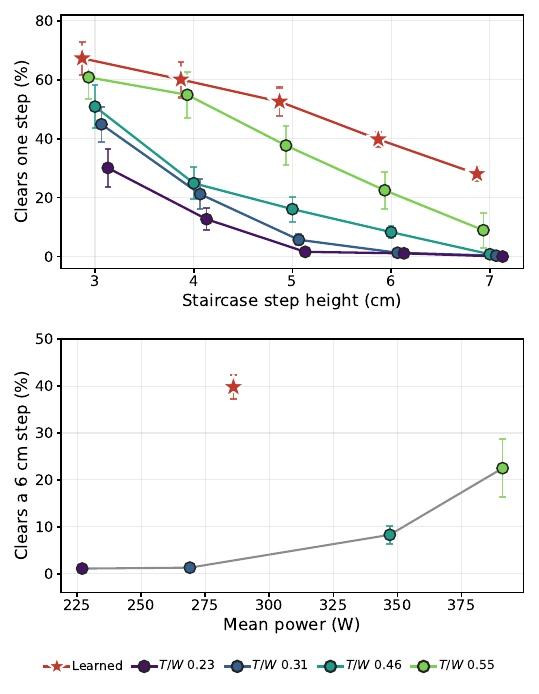}
    \caption{Fixed-thrust comparison. \emph{Top:} single-step clearance against step height for the learned policy and four fixed allocations.
    \emph{Bottom:} clearance of a \SI{6}{\centi\metre} step against mean power.
    Wheels and servos remain learned in all variants. Each point is 2000 episodes and the bars denote one standard error across ten checkpoints. The line joins the fixed-allocation points.}
  \label{fig:accu_ablation}
\end{figure}

We first check whether thrust is needed at all. With the propellers switched off
and only the wheels driving, the robot does not climb a \SI{6}{\centi\metre} step
in any of the evaluated episodes. Flying over the step is the opposite extreme. By
the calibrated power model, hovering in place draws
about 2.5 times the power that the learned policy uses while
climbing, before any forward motion is added.

% Flying over the step is the opposite extreme. Given the full hardware thrust envelope, a policy that leaves the ground rather than climbing draws about 2.4 times the power.

% Flying over the step is the opposite extreme, and it is far less efficient than climbing with thrust assistance. By the calibrated power model, hovering in place alone draws about 2.5 times the power that the learned policy uses while climbing, before any forward motion is added.

Between these two extremes, the question is whether the amount of thrust needs to change during a climb. To isolate thrust allocation, we train four further policies in which wheel and servo control are learned as before while the propeller command is held at one of
four fixed thrust-to-weight ratios (\Cref{fig:accu_ablation}).
The highest of these settings nominally corresponds to \SI{55}{\percent} of body
weight, but because of the simulated thrust ceiling (\Cref{subsec:sysid}) it
reaches only about half of body weight in practice. These controllers can still balance and adapt their wheel motion, but cannot change thrust magnitude during an episode. Each policy is evaluated on
staircases with fixed step heights from \SIrange{3}{7}{\centi\metre}.

\Cref{fig:accu_ablation} shows that the benefit of modulation increases on taller
steps. A sufficiently high fixed allocation is competitive on shallow terrain,
but its clearance falls more rapidly as step height approaches and exceeds the
wheel radius. At \SI{6}{\centi\metre}, the learned policy clears approximately
1.8 times as many episodes while drawing \SI{286}{\watt}, compared with
\SI{391}{\watt} for the best fixed allocation. The approximately
\SI{27}{\percent} lower power is therefore not obtained merely by giving up
climbing capability. No tested fixed allocation reaches both higher clearance and lower power
(\Cref{fig:accu_ablation}, bottom).

\begin{figure*}[t]
  \centering
  \includegraphics[width=\linewidth]{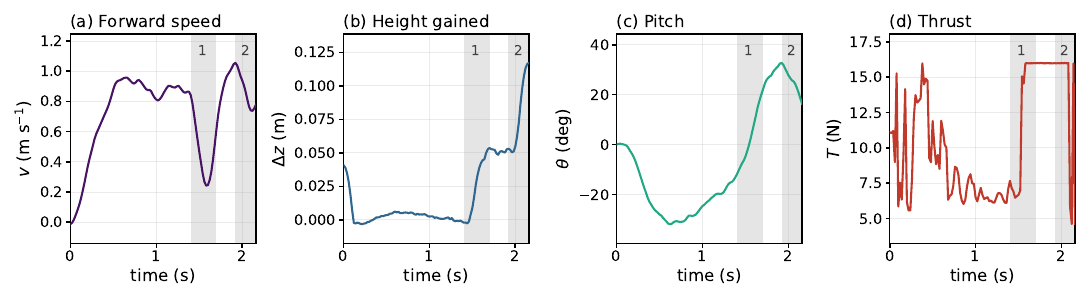}
  % \caption{Thrust and posture through a two-step climb in simulation where shaded
  %   bands mark the step crossings. Forward speed (a) falls as each wheel loads
  %   against a step face and recovers between them. Height (b) rises in
  %   two stages. The body leans back to \SI{-31}{\degree} on the approach and forward
  %   to \SI{33}{\degree} through the crossings (c). Thrust (d) is held near
  %   \SI{7}{\newton} while driving, raised to the simulated thrust ceiling for the climb, and released once the robot is up.}
  \caption{{Thrust and posture through one simulated two-step climb, where shaded bands mark the step crossings. Forward speed (\emph{a}) falls as each wheel loads against a step face and recovers between them. Height (\emph{b}) rises in two stages. The body leans back to \SI{-31}{\degree} on the approach and forward to \SI{33}{\degree} through the crossings (\emph{c}). Thrust (\emph{d}) settles near \SI{7}{\newton} while driving, rises towards the simulated ceiling as each step is crossed, and falls away afterwards.}}
  \label{fig:sim_thrust_modulation}
\end{figure*}

A fixed allocation must provide support at the edge while paying the same propeller cost elsewhere. Continuous allocation can adapt that support to the current state. \Cref{fig:sim_thrust_modulation} shows one climb, where thrust settles near \SI{7}{\newton} while driving and rises only around the crossings.

% \textcolor{red}{These fixed-thrust policies are trained independently rather than
% branched from the learned policy's pre-training checkpoint, so the
% comparison holds the evaluation protocol fixed but not the training
% history. The mode-switching study in the next subsection controls for
% that, since those variants share the starting checkpoint and budget.}

These fixed-thrust policies are trained independently and hence they match the evaluation protocol but not the training history, which the next subsection controls for.

%%%

%%%

\subsection{Continuous Thrust versus Mode Switching}
\label{subsec:mode_switch}

The fixed-allocation comparison leaves open whether a simple switch can recover
the same benefit. We therefore evaluate two-level thrust controllers with the
same learned wheel and servo action space. Each raises thrust when the local
height scan detects a step, which occurs when a ray within \SI{0.105}{\metre} ahead
exceeds the rear rays by \SI{2}{\centi\metre}. The high-thrust state is then held
for \SI{3}{\second}, allowing support to persist while the robot straddles the edge.

Both variants use a high $T/W$ of 0.55, and their low levels are 0.31 and 0.46.
The variants and the learned-thrust comparison share a starting checkpoint,
reward, terrain and training budget. An additional test-time sweep of 18 latch, detection-threshold and high-thrust settings per variant did not improve on the trained configuration.

\begin{table}[t]
\centering
\caption{Learned thrust against mode switching. Clearance is in percent, power is measured at \SI{6}{\centi\metre}, and values after $\pm$ are one standard error across checkpoints.}
\label{tab:mode_switch}
\footnotesize
\setlength{\tabcolsep}{3.1pt}
\begin{tabular}{@{}lrrrr@{}}
\toprule
% & Learned & Switch 0.31 & Switch 0.46 & Fixed 0.55$^\dagger$\\
& Learned & Switch 0.31 & Switch 0.46 & Fixed 0.55\\
\midrule
3\,cm & $67.3\pm5.6$ & $\mathbf{70.7}\pm4.7$ & $62.6\pm6.5$ & $60.9\pm7.3$\\
4\,cm & $60.1\pm6.1$ & $\mathbf{60.4}\pm7.0$ & $54.1\pm6.9$ & $54.9\pm7.7$\\
5\,cm & $\mathbf{52.6}\pm4.8$ & $39.8\pm5.6$ & $42.6\pm3.7$ & $37.7\pm6.7$\\
6\,cm & $\mathbf{39.8}\pm2.5$ & $28.8\pm5.0$ & $25.6\pm3.7$ & $22.5\pm6.2$\\
7\,cm & $\mathbf{28.1}\pm2.4$ & $15.9\pm3.8$ & $13.1\pm2.5$ & $8.9\pm5.9$\\
\midrule
Power (W) & 286 & 264 & 321 & 391\\
\bottomrule
\end{tabular}
% \par\smallskip
% \begin{minipage}{\columnwidth}\footnotesize
% $^\dagger$Trained independently of the learned policy's checkpoint.
% $^\dagger$Trained independently of the learned policy's checkpoint, so its training history is not matched.
% \end{minipage}
\end{table}

The main distinction in \Cref{tab:mode_switch} is capability on taller steps,
not uniformly lower power. On shallow steps, the lower-thrust switch matches the
learned policy within the reported variability. From
\SIrange{5}{7}{\centi\metre}, learned thrust improves clearance over both switches
by roughly 10--15 percentage points, so the tested two-level rules do not
recover the full single-step benefit of continuous allocation. Since both switches reach the same high thrust as the best fixed allocation, the difference lies in how thrust is shaped through the contact rather than in how much is available. 

% \Cref{fig:sim_thrust_modulation} shows that shaping in one episode, where thrust follows the evolving attitude and contact rather than switching between two levels.

The power comparison also exposes the trade-off. The lower-thrust switch uses
less mean power than the learned policy, but clears fewer of the taller steps,
while the higher-thrust switch uses more power and also clears fewer. Among
episodes climbing one \SI{6}{\centi\metre} step, the learned policy uses
\SI{658}{\joule}, compared with \SI{707}{\joule} and \SI{794}{\joule} for the two
switches. Mean power and energy conditioned on a completed climb answer different
questions, since duration and the selection of episodes affect their relationship.

% The observed advantage remains specific to the evaluated rules and single-step metric.

\subsection{Energy Weight and the Cost of Reliability}
\label{subsec:energy_ablation}

To separate the energy objective from differences in initial skill, we fine-tune
five runs from one checkpoint whose pre-training carried no energy penalty,
varying only $w_E\in\{0,2,4,6,8\}$ under an identical budget. Evaluation uses
\SI{6}{\centi\metre} steps.

Energy is compared at a fixed number of steps climbed, because a policy that
simply climbs less would otherwise appear efficient. Conditioned that way, the
penalty does what it is meant to. Mean energy on one-step episodes falls
monotonically from \SI{740}{\joule} at $w_E=0$ to \SI{505}{\joule} at $w_E=8$,
and is \SI{658}{\joule} at the setting we use.

Reliability costs most of what it is going to cost early, and is then stable
until the penalty grows large. Single-step clearance drops from $52.6\pm5.4\%$ at
$w_E=0$ to $39.8\pm2.5\%$ at $w_E=4$, is unchanged through $w_E=6$ within the
reported uncertainty, then collapses to $11.8\pm3.6\%$ at $w_E=8$, where most
episodes never clear a step. We use $w_E=4$, the weight used for the hardware
policy.

\subsection{Why Arrival Quality Matters}
\label{subsec:success_metric}

The terminal criterion distinguishes reaching the neighbourhood of a goal from
completing a usable climb. Rescoring the same trajectories with horizontal
proximity alone inflates success by up to $6.3\times$ across the tested
\SIrange{4}{6}{\centi\metre} steps. Uprightness accounts for most of the
extra accepted states, many of which are arrivals with the robot fallen or still
rotating, rather than balanced on the target platform.

This result supports gating the goal bonus on posture and body rate as well as
position. It is a metric comparison on the same trajectories, not a
separate demonstration that the success gate alone improves learning. Nor does
one accepted terminal state prove indefinite stability, and post-climb behaviour is examined in the following subsection.

\subsection{Hardware Climbing and Transfer}
\label{sec:hardware}

\subsubsection{Primary two-step experiment}

We evaluate two \SI{6}{\centi\metre} steps spaced \SI{40}{\centi\metre} apart,
matching the trained tread and the wheel-radius reference height. Trials start
\SI{90}{\centi\metre} before the first step, with the goal on the upper step. Simulation, in contrast, samples goals at random. The network and deployment interface are held fixed across all ten trials.

Measurement starts when the policy takes control and ends when the target height
is first reached, rather than at operator disarm. Height is measured relative to
the robot's upright standing height, and step clearance additionally requires the
body to be within \SI{25}{\degree} of upright. Height alone can be misleading
because leaning or lifting one wheel also raises the base, and the uprightness
check reduces this ambiguity. Ending at the target height avoids an arbitrary
operator-dependent delay in the energy window. This protocol is not identical to
the simulation's full settled-goal criterion.

\begin{table}[t]
\centering
\caption{Primary hardware experiment. Distance, duration, power and energy refer
to the eight completed climbs, and values after $\pm$ are standard deviations
across trials.}
\label{tab:hardware_primary}
\footnotesize
\begin{tabular}{@{}p{0.62\columnwidth} r@{}}
\toprule
Metric & Measurement\\
\midrule
Both steps cleared & 8/10 trials\\
Distance travelled & \SIrange{1.26}{1.40}{\metre}\\
Duration & \SI{2.95(64)}{\second}\\
Mean electrical power & \SI{275(21)}{\watt}\\
Integrated electrical energy & \SI{823(229)}{\joule}\\
\bottomrule
\end{tabular}
\end{table}

The robot clears both steps in 8 of 10 trials (\Cref{fig:overview}), and the
principal measurements are collected in \Cref{tab:hardware_primary}. In the
representative sequence, forward wheel motion is combined with lift at the edges,
after which the robot balances on the upper step. In trials where the policy
remains active after arrival, upright balancing is observed for up to
\SI{3}{\second}. This balancing occurs after the measurement window closes and is not included in the reported duration or energy.

% This post-climb observation is separate from the timed clearance window.

Power varies less than total energy, indicating that much of the energy
variation accompanies differences in traversal time. The measured mean power
of \SI{275}{\watt} is also close to the reported \SI{286}{\watt} in simulation
at the same step height. The hardware figure is total battery power, while the simulated value covers the propellers and wheels only.

\subsubsection{Additional terrain configurations}
\begin{figure}[t]
  \centering
  \includegraphics[width=\linewidth]{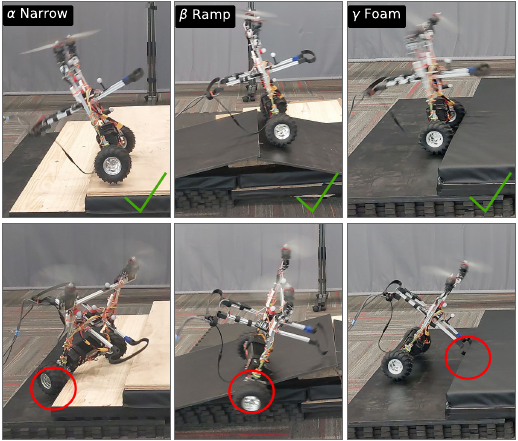}
  \caption{Additional hardware terrains. \emph{Top:} representative traversals on
  ($\alpha$) closely spaced unequal steps, ($\beta$) a ramp crest and
  ($\gamma$) taller steps with a foam upper surface. \emph{Bottom:} asymmetric wheel
  contact, departure from the ramp edge and chassis jamming. Geometry and trial
  outcomes in \Cref{tab:hardware_terrain}.}
  \label{fig:hw_terrain}
\end{figure}

We retain the network for three further configurations, with five trials on each
(\Cref{fig:hw_terrain} and \Cref{tab:hardware_terrain}). The table separates reaching a
surface from settling upright on it, rather than reporting these different
outcomes as a single success rate. Interface parameters are adjusted for each terrain (\Cref{subsec:deployment}).

% \begin{table}[t]
% \centering
% \caption{Additional terrain tests, five trials each. Outcome definitions differ
% by configuration and are stated explicitly.}
% \label{tab:hardware_terrain}
% \footnotesize
% \begin{tabular}{@{}l p{0.40\columnwidth} p{0.44\columnwidth}@{}}
% % \begin{tabular}{@{}>{\raggedright}p{0.22\columnwidth} >{\raggedright}p{0.33\columnwidth} >{\raggedright\arraybackslash}p{0.37\columnwidth}@{}}
% \toprule
%  & Terrain & Observed outcome\\
% \midrule
% $\alpha$ & Steps of 6, 7, 8\,cm; 20\,cm spacing
%  & Third step reached: 4/5; upright on top: 3/5\\
% $\beta$ & Ramp ascent and descent at $12.9^\circ$
%  & Crest crossed and far side descended: 4/5\\
% $\gamma$ & Steps of 8 and 9\,cm; second step is foam
%  & First step cleared: 5/5; upright on second: 2/5\\
% \bottomrule
% \end{tabular}
% \end{table}

The unequal and taller steps show that lifting can carry the wheels beyond the
trained geometry, including steps above the wheel radius. The ramp is also traversed, although the supplied map, and hence the height scan, described the two-step staircase of the primary experiment rather than the ramp. This observation is consistent with a useful role for proprioceptive feedback, but it does not isolate that role or establish insensitivity to terrain-map errors. Likewise, successful runs on foam do not remove contact uncertainty, since deformation changes the support surface under load.

\begin{table}[t]
\centering
\caption{Additional terrain tests, five trials each. Outcome definitions differ
by configuration and are stated explicitly.}
\label{tab:hardware_terrain}
\footnotesize
\renewcommand{\arraystretch}{1.2}
\begin{tabular}{@{}l >{\raggedright}p{0.36\columnwidth} >{\raggedright\arraybackslash}p{0.40\columnwidth}@{}}
\toprule
 & Terrain & Observed outcome\\
\midrule
$\alpha$ Narrow & Steps of 6, 7, 8\,cm\newline with 20\,cm spacing
 & Third step reached\hfill 4/5\newline Upright on top\hfill 3/5\\ \addlinespace[2pt]
$\beta$ Ramp & Two $12.9^\circ$ ramps\newline meeting at a crest
& Crest crossed and\newline far side descended\hfill 4/5\\ \addlinespace[2pt]
$\gamma$ Foam & Steps of 8 and 9\,cm\newline with second of foam
 & First step cleared\hfill 5/5\newline Upright on second\hfill 2/5\\
\bottomrule
\end{tabular}
\end{table}

\subsubsection{Failure mechanisms and remaining limits}
The observed failures point to three limitations rather than a single thrust
shortfall.

% \textbf{Asymmetric contact and heading drift.} When one wheel catches a step while
% the other moves, the robot pivots about the constrained wheel. On a narrow ramp,
% accumulated heading error can carry a wheel off the edge.

% Restoring the wheel differential disabled by the interface (\Cref{subsec:deployment}) is therefore an important deployment change, although the present trials do not quantify its benefit.

\textbf{Asymmetric contact and heading drift.} When one wheel catches a step
while the other moves, the robot pivots about the constrained wheel, and on a
narrow ramp the accumulated heading error can carry a wheel off the edge. The
interface holds the wheel differential at zero (\Cref{subsec:deployment}), so the
hardware trials do not exercise the policy's yaw output. Whether restoring it
would prevent the drift is untested, since steering acts through the same wheel
contact that the caught wheel has already lost.

\textbf{Stalled lean without recovery.} After an unsuccessful approach, the robot
can remain leaned against the step with substantial wheel slip and no renewed
climb. The policy does not back away and retry. Training episodes end once the robot leans beyond \SI{70}{\degree}, so large leans are rarely seen during training. The robot also stalled at smaller leans, however, so this cutoff is only part of the explanation, and the trials show that recovery is missing without identifying a single cause.

\textbf{Mechanical jamming.} The next edge can arrive before the robot has recovered
its attitude, allowing the chassis to wedge against the step face, after which large actuator commands produce little motion. Difficulty therefore depends on the previous step's height and the settling distance available, not on step height alone.

Together these mechanisms account for the unsuccessful trials in \Cref{tab:hardware_primary} and \Cref{tab:hardware_terrain}. These stalled states also cost energy, and the most expensive run is a jam
consuming \SI{1788}{\joule}. Improving energy per useful traversal therefore needs timely detection of unproductive contact and a trained retreat or retry behaviour.

\section{Conclusion and Future Work}
\label{sec:conclusion}

We presented an energy-aware reinforcement learning framework in which a single policy continuously coordinates the wheels, tilt servos and propellers of DoubleBee. Calibrated propulsion costs and mechanically motivated rewards enable thrust-assisted climbing without a discrete locomotion-mode variable. The tested policy improves single-step clearance over both fixed-thrust and mode-switching baselines on taller steps, while hardware trials demonstrate repeated two-step climbs on a physical prototype, along with additional trials on different terrain geometries. Sustained multi-step reliability, onboard perception, heading control, explicit recovery and retraining with actuator models that match the hardware remain the main directions for further development of the system. Stable climbs are rare early in training, and seeding it with priors distilled from the policy's own successful climbs, through self-imitation or adversarial motion priors, could reach them faster than reward shaping alone.

\bibliographystyle{IEEEtran}
\bibliography{references}
\end{document}